\documentclass[authoryear,final,3p,12pt]{elsarticle}

\usepackage{amssymb}
\usepackage{amsmath}
\usepackage{booktabs}
\usepackage{xcolor}
\usepackage{float}
\usepackage{graphics}
\usepackage{enumitem}
\usepackage{subfig}
\usepackage{blkarray, bigstrut}
\usepackage{bm}
\usepackage{multicol,multirow}
 \usepackage{amsthm}
\usepackage{listings}
\theoremstyle{definition}

\usepackage{natbib}
\usepackage{hyperref}
\hypersetup{
citecolor= magenta, 
    colorlinks=true,
    linkcolor=cyan, 
    filecolor=magenta,      
    urlcolor=cyan,
    pdftitle={Overleaf Example},
    pdfpagemode=FullScreen,
    }

\date{}

\begin{document}

\begin{frontmatter}




\title{A machine learning approach to predict university enrolment choices through students' high school background in Italy}

 \author[label1]{Andrea Priulla\corref{cor1}}
 \author[label1,label2]{Alessandro Albano}
  \author[label1]{Nicoletta D'Angelo}
   \author[label1]{Massimo Attanasio}
 \address[label1]{Dipartimento di Scienze Economiche, Aziendali e Statistiche,  Universit\`{a} degli Studi di Palermo, Italy}
 \address[label2]{Sustainable Mobility Center (Centro Nazionale per la Mobilità Sostenibile—CNMS)}
 
\cortext[cor1]{Corresponding author: \texttt{andrea.priulla@unipa.it}}

\begin{abstract}
This paper explores the influence of Italian high school students' proficiency in mathematics and the Italian language on their university enrolment choices, specifically focusing on STEM (Science, Technology, Engineering, and Mathematics) courses. 
We distinguish between students from scientific and humanistic backgrounds in high school, providing valuable insights into their enrolment preferences. Furthermore, we investigate potential gender differences in response to similar previous educational choices and achievements. The study employs gradient boosting methodology, known for its high predicting performance and ability to capture non-linear relationships within data, and adjusts for variables related to the socio-demographic characteristics of the students and their previous educational achievements. Our analysis reveals significant differences in the enrolment choices based on previous high school achievements. The findings shed light on the complex interplay of academic proficiency, gender, and high school background in shaping students' choices regarding university education, with implications for educational policy and future research endeavours.
\end{abstract}

\begin{keyword}
gender gap \sep educational data \sep  university enrolment \sep STEM \sep machine learning \sep gradient boosting 



\end{keyword}

\end{frontmatter}

\section{Introduction}

Since the Second World War, there has been a remarkable expansion of educational opportunities worldwide, leading to an increased demand for education at all levels. This growth was expected to promote the development of democratic principles and reduce educational inequalities \citep{hadjar2009, kromydas2017}. However, despite this significant progress, disparities in educational attainment persist across various social groups.

An important area of focus in international research has been the examination of gender as a primary determinant of educational choices, particularly in higher education. 
For instance, numerous researchers in the US have shown that females generally outperform males in reading test scores, grade completion, and repetition rates at school \citep{stoet2018gender}. Additionally, they exhibit a higher probability of opting for academic educational programs in high school, attending tertiary education, and achieving bachelor's degrees \citep{legewie2012school}.\\
Despite recent advancements in female educational attainment at the secondary and tertiary levels, there has been limited success in reducing gender horizontal segregation, that is, the different orientations of males and females in terms of educational choices worldwide \citep{macarie2015horizontal, cheryan2017some, barone2019nudging,romitogender}. 
Research reveals indeed significant gender differences, particularly along the humanistic-scientific divide, with females being underrepresented in Science, Technology, Engineering, and Mathematics (STEM) or STEM-related fields \citep{cheryan2012understanding, gabay2015gender, tandrayen2021gender}. One notable statistic, derived from OECD countries in 2020, shows females accounted for only 31\% of new enrolments in STEM fields at the bachelor's level \citep{oecd2022}. These patterns are consistently observed across countries, highlighting the existence of structural barriers that perpetuate gender segregation at different educational levels, such as master's (see \cite{priulla2023unveiling} for the Italian case) or doctoral levels \citep{lorz2019gender}.
 
Research aimed to identify individual and contextual factors that lead males and females to make different educational choices \citep{regan2014attitudes}. There has been a particular focus on the lack of female role models \citep{gonzalez2020girls} and the potential discouragement of females from pursuing careers in STEM, stemming from a heavily stereotyped culture and the influence of teachers, peers, and often parents \citep{archer2012science,tey2020teacher,porcu2022estimating}. Further research has highlighted the role of previous educational choices and achievements on the choice to pursue a specific university career \citep{priulla2023does}. In particular, the concept of primary and secondary effects introduced by \cite{boudon1974education} has been adapted to the framework of gender inequalities in education.  According to the literature \citep{hadjar2016education,hadjar2019educational}, the primary effects of gender on educational attainment encompass disparities in performance and achievement between boys/men and girls/women that arise from personal attitudes exhibited by males and females. On the other hand, the secondary effects involve the considerations made by parents, teachers, and students regarding the probability of succeeding for males and females, shaping educational outcomes, influencing decision-making processes, and setting expectations based on gender.
In this work, the direction followed is closer to the themes identified by the primary effects.

Focusing on the Italian case, some insights emerge from a recent report by the Italian National Statistical Institute (ISTAT), shedding light on Italy's considerable disparity in STEM participation compared to other European countries. Specifically, STEM graduates constituted 24.7\% out of the total graduate population in Italy in 2021, trailing behind figures of 26.8\% in France, 27.5\% in Spain, and 32.2\% in Germany. Notably, the data highlights that females show significantly lower participation than men towards STEM disciplines: out of 100 women earning a tertiary degree, only 15 obtained it in a STEM field, compared to 33 of males \citep{ISTAT2023}.

In this respect, the international literature has extensively highlighted the importance of adopting models capable of considering the intricate structure of educational differences \citep{borman2010schools,giambona2018school}. 

Many classical statistical models have been employed in literature to study these differences.
However, in the last decade,  
widespread adoption of flexible and non-parametric machine learning models has revolutionized various scientific domains. These models excel in capturing intricate and non-linear relationships within data, demonstrating adaptability to diverse data distributions without relying on specific assumptions. They also can automatically manage relevant features during training, exhibit robustness to overfitting when appropriately tuned, showcase high predictive performance, and efficiently handle large datasets.
In recent years, the application of machine learning methods has also extended to the realm of education, where they play a pivotal role in analyzing educational data to offer valuable insights for enhancing the learning process 
\citep{fernandes2019educational, yaugci2022educational, suzuki2022prediction}.

Our research aims to investigate the relationship between high school backgrounds and university enrolment choices in Italy. 
Our primary focus lies in discerning the influence of previous educational achievement on the choices made by males and females concerning university enrolment and, subsequently, the choice of a specific field of study at university.
Moreover, we focus on students who excel in both mathematics and Italian tests to identify further potential differences in their future academic choices. This investigation will provide a further understanding of the factors driving academic success and contribute to the ongoing discourse on gender-related differences within educational contexts.
To this aim, we use longitudinal micro-data from two Italian administrative sources, namely the Ministry of University and Research (MUR) and INVALSI. This integrated database allows to follow students from the fifth and last year of high school up to the end of their first year of university career.

The paper is structured as follows. Section \ref{sec:lit} provides a literature review on the topic of gender differences in STEM. Section \ref{sec:data} illustrates the data. Section \ref{sec:methods} contains the methods employed. 
Section \ref{sec:exploratory} provides an exploratory analysis, and Section \ref{sec:results} delves into the results of the gradient boosting procedure.
The paper ends with some conclusions in Section \ref{sec:discussion}.

\section{Literature review}\label{sec:lit}



This section is devoted to the review of the international literature on the gender gap in STEM, with particular reference to the Italian case.

Most of the worldwide literature on such topics generally focuses on student performance in mathematics. Research has indeed shown that, among a variety of factors, overall academic achievement and proficiency in maths are crucial predictors of the choice to enroll in a STEM program at university. In detail, the gender gap in tertiary education arises from factors that manifest before this stage and shape students' interest towards a specific field.
A recent strand of international literature has highlighted how students' educational choices conform to a gender divide as early as primary school \citep{bian2017gender,makarova2019gender}. 

From a cultural point of view, a stereotype is that females possess an inherent inclination towards educational paths emphasizing humanistic and caring disciplines. Over the years, various theories have been explored to elucidate these differences in educational preferences. For instance, \cite{sherman1980mathematics} proposed that family dynamics, school environments, and teachers' attitudes significantly shape the attitudes of males and females towards specific subjects and skills, influencing their educational decisions.
Further research has delved into the examination of implicit gender stereotypes concerning individuals' mathematical identities \citep{cvencek2011math} and has investigated the positive correlation between STEM identity and various outcomes, including persistence and career goals \citep{mcnally2020gender}. 
According to balanced identity theory, if a girl holds stereotypes associating math and science with male proficiency, it is anticipated that she will disassociate herself from these domains and actively avoid pursuing advanced courses and careers within them.
On the other hand, \cite{correll2001gender} argues that gender differences in mathematics are not solely responsible for the significant imbalances between the numbers of males and females entering fields requiring advanced mathematical competence. Cultural beliefs about gender and mathematics influence the choices of males and females toward STEM careers differently. The author suggests that some individuals may come to personally believe that males are inherently better at math, even though females are less likely than males to hold stereotypical views about mathematics. As a result, if a girl believes that males excel in math, she may perceive her mathematical competence as contradictory with her female gender identity, leading to self-doubt and decreased interest in careers requiring high mathematical competence. 
The perception that others hold these gendered beliefs about mathematics can lead to biased self-assessments and reduced performance. In this sense, since males tend to overestimate their mathematical competence relative to females, they are also more likely to pursue activities that pave the way for STEM careers.

While these results mainly come from research conducted in the United States, they also seem adaptable to European social structures.
European literature has indeed widely addressed the topic of STEM engagement, analysing the factors affecting the choice to pursue a STEM career \citep{smith2011there,regan2014attitudes}. In detail, research has shown that despite performing well in mathematics and literacy, males and females attribute different importance to their prior educational achievements when it comes to choosing their academic path. The influence varies by subject, as evidenced by studies. For instance, in the UK, boys tend to be more influenced by their comparative advantage in English and maths when making STEM-related choices compared to girls \citep{delaney2020effect}
Moreover, in an investigation of Israeli high school course preferences, girls were found to be more responsive to prior grades in biology and chemistry, while boys exhibited stronger reactions to grades in computer science and physics \citep{friedman2016gender}. 

\cite{barone2019nudging} presents another perspective concerning Italy, underscoring the lack of accurate information in high schools regarding the long-term job prospects associated with specific degree programs. This absence of information on economic rewards and career opportunities prompts students to base their choices solely on their preferred subjects or ``dream'' occupations, often influenced by gender stereotypes. 
From another perspective, the rational choice theory suggests individuals tend to favor educational options that enhance their chances of success \citep{barone2020gender}. This theory posits that gender differentiation results from socialization processes and rational choice factors \citep{gabay2015gender}. According to this theory, students with a more career-oriented mindset are less likely to enrol in non-STEM programs. 

In Italy, as in many other European countries, the educational careers of the students are largely influenced by the choice of a specific curriculum in high school. 
The Italian high school system can be seen as a hierarchical tripartite structure. The system comprises ``licei'', with a focus on humanities and sciences, serving as traditional institutions preparing students for potential university enrollment; technical schools and their diverse tracks are positioned as an intermediate choice, bridging academic and vocational pathways; vocational schools aim to equip students for direct entry into the labor market, providing training for various low-ability jobs.
It is important to notice that since this choice occurs at the age of 13 in Italy, it is often not an independent choice made by students but is significantly influenced by their parents \citep{contini2016between}.

Despite universalist principles, European educational systems, including the Italian one, have indeed historically been organized to segregate students based on social class and gender. This structuring has compelled families to enrol their children in distinct educational institutions, each conforming to specific social, educational, and gender-based stratification and norms \citep{salmieri2020gender}. 
The academic achievements and gender-specific subject preferences observed at a particular educational stage, especially in a highly stratified high school system like the Italian one, serve as reflections of previous achievements and preferences in earlier phases of the students' educational journey \citep{salmieri2022students}.
In this regard, \cite{contini2023gender} analysed the effect of previous achievement in mathematics and Italian on the choice of the high school curriculum in Italy, showing that female students require stronger prior signals of mathematical ability to choose STEM fields.
Then, it is not surprising that in Italy, the gender distribution across different educational paths is notably heterogeneous. Females constitute broadly 70\% in the classical curriculum, slightly less than 50\% in the traditional scientific curriculum, and only 32\% in the applied sciences one. The latter was established in the latest reform of the high school education system in Italy in 2010, aiming to provide extensive training in studies of scientific and technological fields. The distinction from the traditional scientific high school lies in the replacement of Latin language studies with more hours devoted to mathematics and other scientific subjects.
In a recent paper, \cite{priulla2023does} has emphasized that attending the applied sciences track, which means attending more scientific-related classes in high school, positively influences the probability of enrolment in STEM. 

\section{Data}\label{sec:data}

The dataset used for the analysis is built by linking two distinct administrative national sources: 
\begin{itemize}
\item \textbf{INV-S}: micro-data sourced from the Italian National Institute for the Evaluation of the Education and Training System (INVALSI). INVALSI, operating as a research institution with legal status, conducts nationwide standardized computer-based tests to assess the overall quality of the Italian education system across different types of schools. These tests
are administered annually to students at five educational levels (grades 2, 5, 8, 10, and 13) to evaluate their proficiency in mathematics, Italian language skills, and, since 2018, English reading and listening skills. In this paper, we consider the maths and Italian language tests administered at grade 13.
In addition to the test scores, INVALSI also collects information about socio-demographic status, family background, and further indicators of previous academic performance.
\item \textbf{ANS-U}: micro-level longitudinal data from the Italian National Archive of University Students (ANS) \citep{mobysu2017}. This comprehensive database contains detailed information about the university pathways of all students enrolled in Italian universities between 2010 and 2020. The database provides a comprehensive record encompassing their high school background and subsequent university career.

\end{itemize}

The linkage of these databases allows investigation of i) the transition from high school to university at the individual level and ii) the relationship between student performance in high school and university outcomes. 
We have access to data about all the students enrolled on the last year of high school in Italy in 2018/19, and we have information about those enrolled at university in the subsequent academic year, i.e. 2019/20.  
However, we are unable to differentiate between students who did not enrol in university and those who chose to enrol at university abroad.

In this paper, we focus on the subset of students attending humanistic and scientific curricula in high school. As for the latter, a further specification is adopted to distinguish the traditional scientific and the applied science tracks within the scientific curriculum. We decided to consider this specification following the results of \cite{priulla2023does} that highlighted differences in terms of academic outcomes of the students attending the two scientific tracks in Italy. We consider the following set of covariates in our analysis: 
\begin{itemize}
    \item \textbf{HS macroregion}: the macroregional location of the high school attended by the student
\begin{itemize}
    \item {South \& Islands} (Abruzzo, Basilicata, Calabria, Campania, Molise, Apulia, Sardinia, Sicily)
    \item {Center} (Latium, Marche, Tuscany, Umbria)
    \item {North} (Emilia Romagna, Friuli Venezia Giulia, Liguria, Lombardy, Piedmont, Trentino Alto Adige, Veneto)
\end{itemize}

    \item \textbf{HS SES}: The overall socio-economic status of the school\footnote{The OECD PISA surveys and INVALSI tests use the ESCS index (Index of Economic, Social, and Cultural Status) that synthetically defines the socio-economic and cultural status of the students' families. A negative (positive) value of the index indicates a lower (higher) SES than the Italian average.} 
    \item \textbf{Public/Private}: a variable indicating whether the student attended a public or private high school. 
    \item \textbf{Gender}
    \item \textbf{HS curriculum}: the type of high school curriculum attended by the student. We consider two curricula: the humanistic and the scientific curriculum, divided into traditional and applied sciences tracks.
    \item \textbf{INVALSI math score}:  the scores\footnote{INVALSI, similarly to PISA tests, reports all the scores using the Rasch metric, setting the average for Italy at 200 and the standard deviation at 40.} in the INVALSI math test at grade 13. 
    \item \textbf{INVALSI Italian score}: the scores in the INVALSI Italian test at grade 13.

\end{itemize}

\section{Methods}\label{sec:methods}

This section introduces Gradient Boosting (GBM) \citep{friedman2000additive, friedman2001greedy, friedman2002stochastic}, a machine learning (ML) model employed to predict a response variable based on a set of covariates. In general, ML techniques provide a powerful framework for the analysis of high-dimensional datasets by modelling complex relationships, often encountered in modern data with many variables, cases, and potentially nonlinear effects. The impact of ML methods is still limited in the field of educational sciences but continuously growing as larger and more complex datasets become available \citep{hilbert2021machine}.


GBM is an ensemble ML model that sequentially combines the predictions of multiple weak learners, which are typically decision trees. The goal is to find some function $\hat{F}(\mathbf{x})$ that best approximates the output variable $y$ from the values of input variables $\mathbf{x}$. This is formalized by introducing a loss function $L(y, \hat{F}(\mathbf{x}))$ and minimizing its expectation:
\[
\hat{F} = \underset{F}{\arg\min}\, \mathbb{E}_{\mathbf{x},y}[L(y, F(\mathbf{x}))].
\]

The GBM works seeking an approximation \(\hat{F}(x)\) in the form of a weighted sum of \(B\) functions \(h_m(x)\) from some weak learners:
\[
\hat{F}(x) = \sum_{b=1}^{B} \alpha_b h_b(x),
\]

where $b$ is the iteration index. The process begins with a simple model, often a weak learner like a shallow decision tree, making predictions on the training data, $F_0(x)$.  At each iteration, it fits a weak learner to the negative gradient of the loss function for the current model's predictions:

$$
F_{b}(x) = F_{b-1}(x) + \left(\arg \min_{h_{b}} \left[\sum_{i=1}^{n} L(y_{i},F_{b-1}(x_{i})+h_{b}(x_{i}))\right]\right)(x)
$$

where $n$ is the number of observations in the dataset, and $L$ is a differentiable convex loss function that measures the difference between the true values \(y_j\) and the predicted values \(F_{b-1}(x) + h_b(x_i)\).\\
Given the computational challenges in selecting the optimal function \(h_b\) for an arbitrary loss function \(L\), the steepest descent is employed via functional gradient descent. Iteratively updating the model \(F_{b-1}(x)\) toward a local loss function minimum, the adjustment is made by a small distance \(\alpha\) to maintain a valid linear approximation:
\[
F_{b}(x) = F_{b-1}(x) - \alpha \sum_{i=1}^{n} \nabla_{F_{b-1}} L(y_{i}, F_{b-1}(x_{i})).
\]
This iterative process refines predictions based on ensemble errors, ensuring \(L(y_{i}, F_{b}(x_{i})) \leq L(y_{i}, F_{b-1}(x_{i}))\). \\
The GBM has proven to be robust to overfitting through techniques like shrinkage and the use of shallow trees and flexible to handle different types of data, capturing complex relationships and non-linear patterns.
Popular implementations of GBM include the Gradient Boosting Machine (GBM), XGBoost, LightGBM, and CatBoost, each offering optimizations and enhancements to the original GB algorithm. The \texttt{gbm} package in \texttt{R} of the algorithm \citep{ridgeway2007generalized} was chosen for its popularity and reliability.

\subsection{Interpretability methods}
The discussion now delves into various tools designed to interpret the model predictions effectively.  It is essential to note that beyond their predictive capacity, these models serve a dual purpose in interpretation. These tools are instrumental in identifying the most influential factors and elucidating their impact on the response variable. Through this interpretative lens, the models contribute valuable insights into the determinants that significantly shape students' enrollment choices.

Understanding the inner workings of a ML model and interpreting its predictions are critical aspects of model deployment and decision-making \citep{molnar2020interpretable}. In this section, we delve into three essential model interpretability tools: namely, the relative influence of predictors, the Accumulated Local Effects (ALE) and Multi-dimensional Partial Dependence Plots (PDP). These techniques offer valuable insights into the importance and impact of individual covariates on a model's predictions, aiding in the interpretation and selection of relevant variables.\\
The relative influence of a covariate in a GBM model is a measure indicating the relative importance of each variable in training the model. The relative influence of a covariate is calculated by taking each covariate's contribution for each tree in the model and calculating the relative contribution of the corresponding covariate to the model. Specifically, it is a reduction attributable to each variable in
reducing the loss function in predicting the gradient on each iteration. The relative influence of a covariate is usually normalized so that the sum of the relative influences of all covariates is equal to one.
The relative influence of a covariate can be used to rank the individual variables based on their importance in the model. The ranking can be used for covariate selection, where the most important covariates are selected for the final model. However, it is important to note that the relative influence of a covariate does not provide any explanations about how the variable actually affects the response, for this aim other methods need to be used.

Among others, the PDP \citep{friedman2001greedy} stands as the most famous visualization tool to investigate how one or more covariates collectively shape the predictions of a ML model. PDP operates by marginalizing the model's predictions over the distribution of the covariates in set $X_C$, the variables in which we hold no specific interest. This allows the function to reveal the relationship between the covariates in set $x_S$, the variables of interest, and the predicted outcome. The partial function $\hat{F}_S$ is obtained by computing average pointwise predictions for a grid  of
$x_S$ values, within the training data:
\[
\hat{F}_S(x_S) = \frac{1}{n} \sum_{i=1}^{n} \hat{F}(x_S,x_{i,C}).
\]
This partial function yields insights into the average marginal effect on predictions for a grid of values of covariates $x_S$, while $x_{i,C}$ are the actual covariate values in the training set.\\
Unfortunately, the correlation between covariates in $x_C$ and $x_S$ introduces complexity, potentially leading to heavy interpolation, i.e. including data points that are highly improbable to observe. This correlation challenges the accurate reflection of the isolated impact of covariates in $x_S$ due to their interdependence with covariates in $x_C$, as pointed out in the literature \citep{apley2020visualizing}. For this reason, the use of one-dimensional PDP plots is generally discouraged.\\
When more than one covariate is included in $x_S$, multi-dimensional PDP plots are obtained.  Multiple PDPs can overcome one-dimensional PDP limitations by visualizing the joint effects of multiple covariates on the predicted outcome. This provides a comprehensive understanding of covariate interactions and their influence on the model's predictions. Furthermore, to enhance the reliability of interpretations and mitigate the risk of extrapolation, the estimated response can be confined within the convex hull defined by the training values of relevant variables, to avoid interpreting the PDP outside the observed data boundaries. This approach ensures robust insights grounded in the empirical distribution of the training dataset, contributing to a more accurate understanding of relationships between variables and predicted responses.
While Multiple PDPs can effectively address the challenge of visualizing the collective effect of multiple predictors on the response, limitations arise with continuous predictors, allowing visualization for only two variables at a time. 

Additionally, the emphasis often lies in isolating and emphasizing the individual influence of each predictor. In these scenarios, ALE \citep{apley2020visualizing} emerges as a valuable tool.  ALE plots provide a faster and unbiased alternative to one-dimensional PDP and can be utilized to interpret the effects of individual covariates on a model's predictions. For a single numerical covariate, the ALE value can be interpreted as the main effect of the covariate at a certain value compared to the average prediction of the data.\\
To estimate local effects, we divide the covariate $X_j$ into $k$ intervals driven by percentiles and compute the differences in the predictions. More specifically, for each $j \in \{1,2, \dots, d\}$, let 
$N_j(k) = (z_{k-1,j}, z_{k,j}] : k = 1, 2, \ldots, K$  be a partition of the sample range of  $\{x_{i,j} : i = 1, 2, \ldots, n\}$ into $K$ intervals. Here we denote \(z_{k,j}\) as the \(k\)-th quantile of the empirical distribution of \(\{x_{i,j} : i = 1, 2, \ldots, n\}\) with \(z_{0,j}\) chosen just below the smallest observation, and \(z_{K,j}\) chosen as the largest one.\\
For \(k = 1, 2, \ldots, K\), let \(n_j(k)\) denote the number of observations that fall into the \(k\)-th interval \(N_j(k)\), so that \(\sum_{k=1}^{K} n_j(k) = n\). For a particular value $x$ of the explanatory variable $X_j$, let \(k_j(x)\) denote the index of the interval into which \(x\) falls, i.e., \(x \in (z_{k_j(x)-1,j}, z_{k_j(x),j}]\).

To estimate the main ALE $(F_{j,ALE}(\cdot))$ of a predictor $(X_j)$, we first compute an estimate of the uncentered effect \(g_{j,ALE}(\cdot)\):
$$
\hat{g}_{j,ALE}(x) = \frac{1}{n_j(x)} \sum_{k=1}^{n_j(x)} \sum_{i: x_{i,j} \in N_j(k)} [F(z_{k,j}, x_{i,\backslash j}) - F(z_{k-1,j}, x_{i,\backslash j})],
$$
for each \(x \in (z_{0,j}, z_{K,j}]\). In the preceding, \(x_{i,\backslash J} = (x_{i,j} : j = 1, 2, \ldots, d; j \notin J)\) denote the \(i\)-th observation of the subsets of predictors \(X_{\backslash J}\). 
The uncentered effect of a covariate value that lies in a certain interval is the sum of the effects of all the previous intervals. This average in the interval is covered by the term Local in the name ALE. The left sum symbol means that we accumulate the average effects across all intervals. The (uncentered) ALE of a covariate value that lies, for example, in the third interval, is the sum of the effects of the first, second and third intervals. The word ``accumulated'' in ALE reflects this. 
The ALE main effect estimator \(\hat{F}_{j,ALE}(\cdot)\) is then obtained by centering $\hat{g}_{j,ALE}(x$ that the mean effect is zero.
\[
\hat{F}_{j,ALE}(x) = \hat{g}_{j,ALE}(x) - \frac{1}{n} \sum_{i=1}^{n} \hat{g}_{j,ALE}(x_{i,j}) .
\]
One-dimensional ALE plots are not defeated in the presence of correlated predictors, unlike marginal plots \citep{gromping2020model}. The reason is that they analyze differences in predictions accumulating the average effects over a predictor's range instead of the average of the predictions themselves. This cumulative approach helps mitigate issues arising from predictor correlation, providing a more stable and accurate representation of a predictor's impact on the model's predictions.

\section{Exploratory analysis}\label{sec:exploratory}
In this section, we present an exploratory analysis of university enrolment choices made by female and male students, exploring the association between factors related to students' previous educational attainment and their university careers.

In Table \ref{tab:tabellone}, we describe students enrolled on the fifth year of high school in Italy in 2018/19, giving a broad overview of how university enrolment choices of males and females differ according to the socio-demographic characteristics of the students and their high school careers.
To aid interpretation, INVALSI test scores have been reported in Table \ref{tab:tabellone} following the proficiency levels directly supplied by INVALSI, which range from 1 to 5. Students with a proficiency level below 3 are classified as low-performing, while those with a proficiency level of 5 are considered high-performing. Furthermore, the school SES has been classified using the quartiles of its distribution, with the 1st quartile representing low SES and the 4th quartile representing high SES. Throughout the remainder of the paper, these variables will be treated as continuous.

As anticipated, female students, on average, are more likely than males to enrol in university but less inclined to choose STEM programs. Further results reveal significant gender differences in enrolment choices. 
The well-known gap between the northern and southern Italian regions reemerges in students' enrolment choices. Specifically, the percentage of non-enrolled students increases from North to South, with a slightly wider gap observed among males. Interestingly, students from the North, especially males, show a higher inclination towards choosing STEM degree programs than their counterparts from the South. The overall socio-economic status of the school proves to be associated with university enrolment. Specifically, students coming from schools with an overall higher SES are more likely to enrol in non-STEM programs. This is particularly evident among male students, as female ones are generally more likely to opt for non-STEM programs.
Regarding the type of high school attended, students from public schools are more likely to enrol in university, displaying a greater preference for STEM degree programs compared to their peers from private schools.
As for the high school curriculum, the results show that students with humanistic backgrounds show minor differences in their choices, with a similar percentage of both males and females preferring non-STEM programs. On the other hand, females attending both scientific curricula are significantly less likely to enrol in STEM programs than their male counterparts.
Finally, it is possible to observe that, as the INVALSI math and Italian scores increase, the percentage of students not enrolling at university decreases. In particular, a bad performance in maths tests appears to be more associated with males' choice not to enrol at university. Conversely, high performance in mathematics and Italian are associated with a higher probability of enrollment in STEM and non-STEM programs, respectively. This pattern appears to vary based on gender: high mathematics proficiency has a more pronounced impact on males' choice to enrol in STEM programs, while a higher score in Italian appears to be more associated with females' decision to pursue non-STEM programs.\\
However, it's crucial to acknowledge that these associations are marginal and do not account for the joint influence of all variables.

\begin{table}[h]
  \centering
  \caption{Student characteristics according to gender and enrolment choice. Cohort of students enrolled on the fifth year of high school in Italy in 2018/19.}
  \renewcommand{\arraystretch}{1.3}
    \setlength{\tabcolsep}{6pt}
      \resizebox{\columnwidth}{!}{
\begin{tabular}{cc|ccc|c|ccc|c}
    \hline
          &       & \multicolumn{4}{c|}{\textbf{F}} & \multicolumn{4}{c}{\textbf{M}} \\
    \hline
    \multicolumn{1}{c|}{\textbf{Variable}} & \textbf{Category} & \textbf{Not enr.} & \textbf{Non-STEM} & \textbf{STEM} & \textbf{Total} & \textbf{Not enr.} & \textbf{Non-STEM} & \textbf{STEM} & \textbf{Total} \\
    \hline
    \multicolumn{1}{c|}{\multirow{3}[2]{*}{\textbf{HS macroregion}}} & \textbf{North} & \textit{13,7} & \textit{46,9} & \textit{39,4} & 25286 & \textit{14,3} & \textit{34,1} & \textit{51,5} & 26559 \\
    \multicolumn{1}{c|}{} & \textbf{Center} & \textit{14,4} & \textit{47,8} & \textit{37,8} & 14193 & \textit{18,7} & \textit{35,9} & \textit{45,5} & 15570 \\
    \multicolumn{1}{c|}{} & \textbf{South \& Islands} & \textit{16,3} & \textit{48,9} & \textit{34,8} & 30979 & \textit{19,5} & \textit{35,8} & \textit{44,7} & 29061 \\
    \hline
    \multicolumn{1}{c|}{\multirow{3}[2]{*}{\textbf{HS SES}}} & \textbf{Low} & \textit{19,9} & \textit{43,4} & \textit{36,7} & 11390 & \textit{22,9} & \textit{29,6} & \textit{47,5} & 12030 \\
    \multicolumn{1}{c|}{} & \textbf{Medium} & \textit{14,5} & \textit{46,7} & \textit{38,8} & 42695 & \textit{16,8} & \textit{33,7} & \textit{49,5} & 43537 \\
    \multicolumn{1}{c|}{} & \textbf{High} & \textit{12,8} & \textit{54,6} & \textit{32,5} & 16373 & \textit{14,6} & \textit{43,7} & \textit{41,7} & 15623 \\
    \hline
    \multicolumn{1}{c|}{\multirow{2}[2]{*}{\textbf{HS type}}} & \textbf{Private} & \textit{26,7} & \textit{47,7} & \textit{25,7} & 4331  & \textit{30,2} & \textit{37,4} & \textit{32,4} & 6268 \\
    \multicolumn{1}{c|}{} & \textbf{Public} & \textit{14,2} & \textit{48,0} & \textit{37,8} & 66127 & \textit{16,1} & \textit{35,0} & \textit{48,9} & 64922 \\
    \hline
    \multicolumn{1}{c|}{\multirow{3}[2]{*}{\textbf{HS curriculum}}} & \textbf{Humanistic} & \textit{12,5} & \textit{63,5} & \textit{24,0} & 19904 & \textit{15,3} & \textit{61,6} & \textit{23,0} & 9033 \\
    \multicolumn{1}{c|}{} & \textbf{Trad. scientific} & \textit{15,5} & \textit{44,0} & \textit{40,5} & 40922 & \textit{17,0} & \textit{34,2} & \textit{48,8} & 42032 \\
    \multicolumn{1}{c|}{} & \textbf{Applied sciences} & \textit{17,9} & \textit{33,0} & \textit{49,0} & 9632  & \textit{18,9} & \textit{25,5} & \textit{55,6} & 20125 \\
    \hline
    \multicolumn{1}{c|}{\multirow{3}[2]{*}{\textbf{INVALSI Math score}}} & \textbf{Low} & \textit{27,3} & \textit{51,3} & \textit{21,4} & 13554 & \textit{36,3} & \textit{39,6} & \textit{24,0} & 9307 \\
    \multicolumn{1}{c|}{} & \textbf{Medium} & \textit{15,2} & \textit{50,4} & \textit{34,4} & 30564 & \textit{21,3} & \textit{39,6} & \textit{39,1} & 25467 \\
    \multicolumn{1}{c|}{} & \textbf{High} & \textit{8,4} & \textit{43,5} & \textit{48,1} & 26340 & \textit{9,8} & \textit{31,0} & \textit{59,2} & 36416 \\
    \hline
    \multicolumn{1}{c|}{\multirow{3}[2]{*}{\textbf{INVALSI Italian score}}} & \textbf{Low} & \textit{31,9} & \textit{40,6} & \textit{27,5} & 8497  & \textit{35,1} & \textit{30,5} & \textit{34,3} & 10553 \\
    \multicolumn{1}{c|}{} & \textbf{Medium} & \textit{14,4} & \textit{47,6} & \textit{37,9} & 43378 & \textit{16,8} & \textit{35,1} & \textit{48,1} & 42649 \\
    \multicolumn{1}{c|}{} & \textbf{High} & \textit{8,5} & \textit{52,2} & \textit{39,3} & 18583 & \textit{8,2} & \textit{38,2} & \textit{53,6} & 17988 \\
    \hline
    \multicolumn{2}{c|}{\textbf{Total}} & \textit{\textbf{15,0}} & \textit{\textbf{48,0}} & \textit{\textbf{37,0}} & \textbf{70458} & \textit{\textbf{17,4}} & \textit{\textbf{35,2}} & \textit{\textbf{47,4}} & \textbf{71190} \\
    \hline
    \end{tabular}}
  \label{tab:tabellone}%
\end{table}%

The last results obtained in Table \ref{tab:tabellone} regarding the effect of high school career on enrolment choices were reasonably expected. Nevertheless, it is necessary to consider an intersectional approach to have a deeper understanding of the gender differences characterizing the transition from high school to university. 

\begin{figure}[h]
    \centering
    \includegraphics[width=\textwidth]{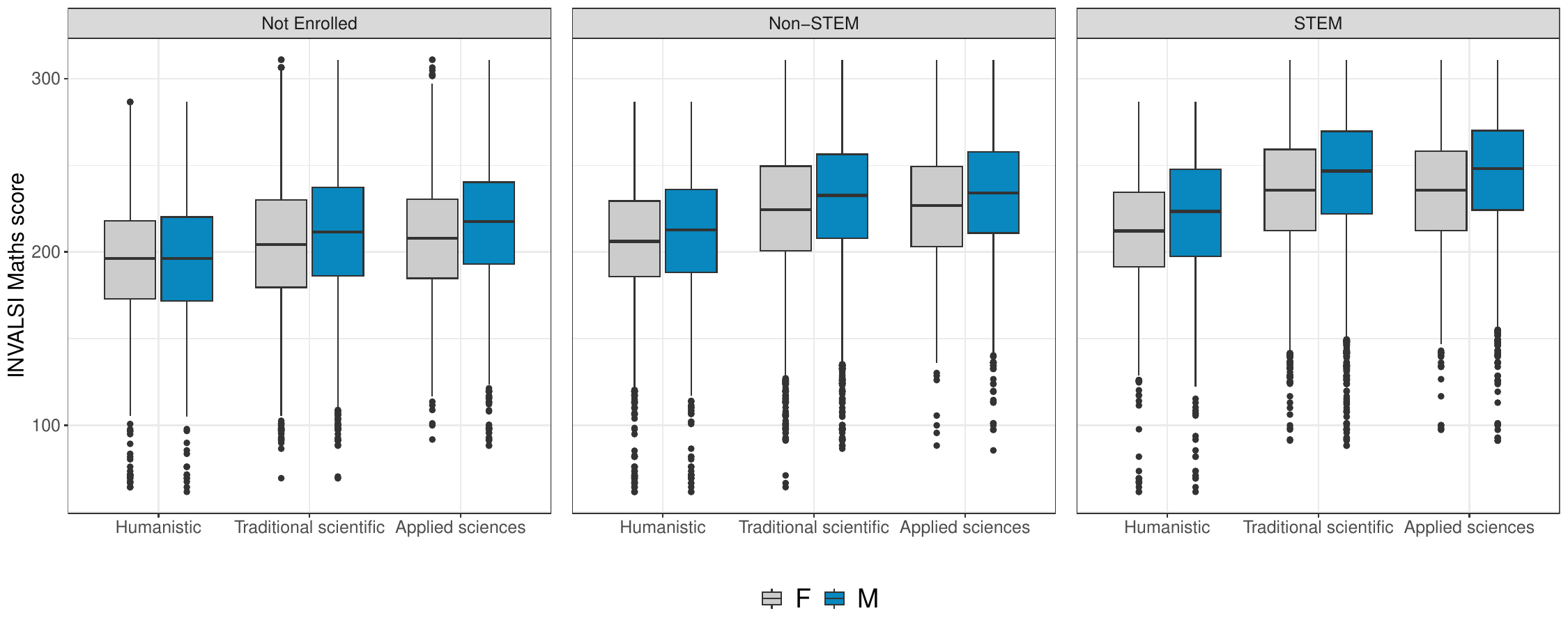}
    \caption{Distributions of maths scores according to gender, the high school curriculum, and enrolment choice.}
        \label{fig:fig1}
\end{figure}

\begin{figure}[htbp!]
    \centering
    \includegraphics[width=\textwidth]{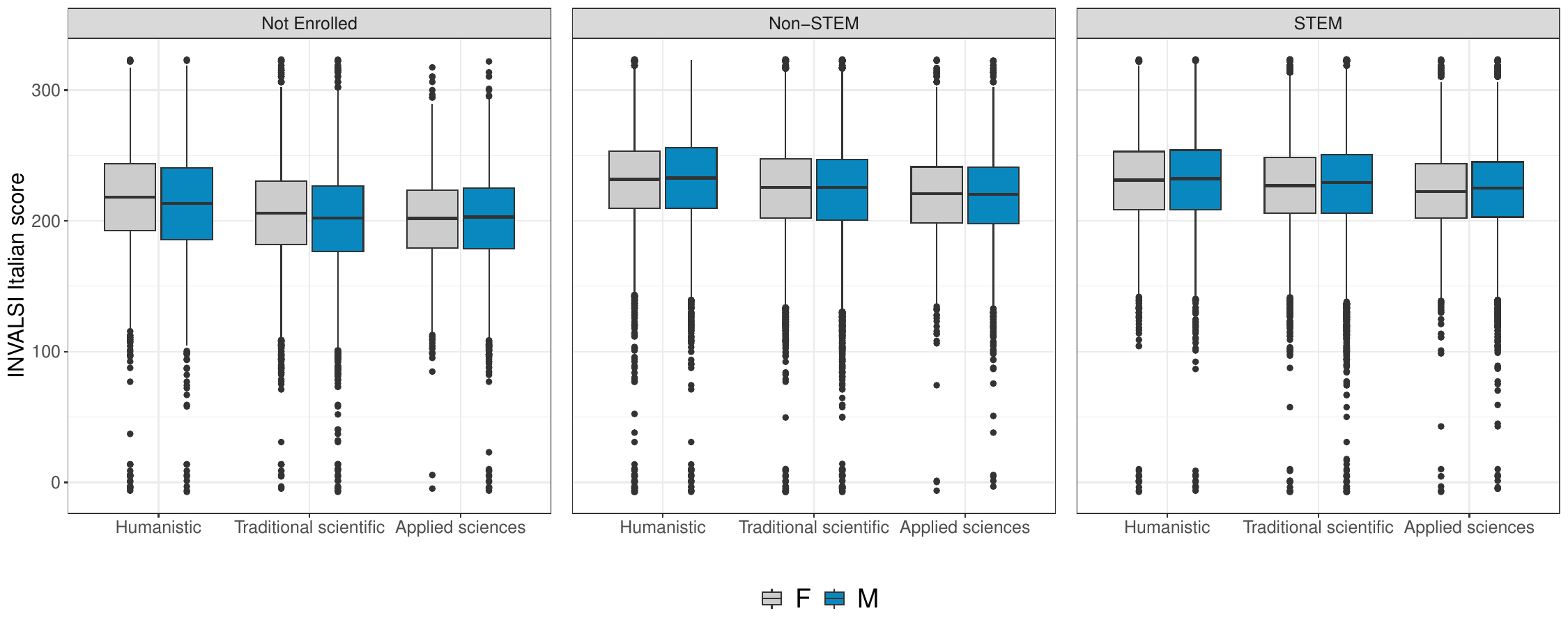}
    \caption{Distributions of Italian scores according to gender, the high school curriculum, and enrolment choice.}
        \label{fig:fig2}
\end{figure}

In Figures \ref{fig:fig1} and \ref{fig:fig2}, we show, respectively, the distribution of maths and Italian scores according to gender, the high school curriculum attended, and the enrolment choice. 
Regarding maths scores (Figure \ref{fig:fig1}), students from scientific high schools generally exhibit better performance. This can be easily explained since humanistic high schools allocate fewer hours to science-related subjects. Moreover, the distribution of scores differs based on enrolment choices: students who opt for STEM programs consistently achieve higher scores in mathematics tests compared to others, but the gap appears to be more pronounced in scientific high schools, suggesting greater confidence in their mathematical abilities among those choosing to pursue STEM degrees. Additionally, it is observable that the gap between males and females is more evident in scientific high schools. In contrast to what is observed in humanistic high schools, male students demonstrate higher performance, even among those who opt not to pursue university enrollment.

Regarding the Italian scores (Figure \ref{fig:fig2}), as expected, students attending the humanistic curriculum perform better than their scientific peers. Nevertheless, differently from what we observed for the maths scores, it seems the Italian scores are still associated with the choice to enrol at university, but not with the choice of the degree program. In fact, students who did not enrol at university show a worse performance in Italian, especially those attending scientific curricula. Moreover, the gap favouring females is slightly more evident among not-enrolled students from humanistic backgrounds.


\begin{figure}[hbtp!]
    \centering
    \includegraphics[width=.95\textwidth]{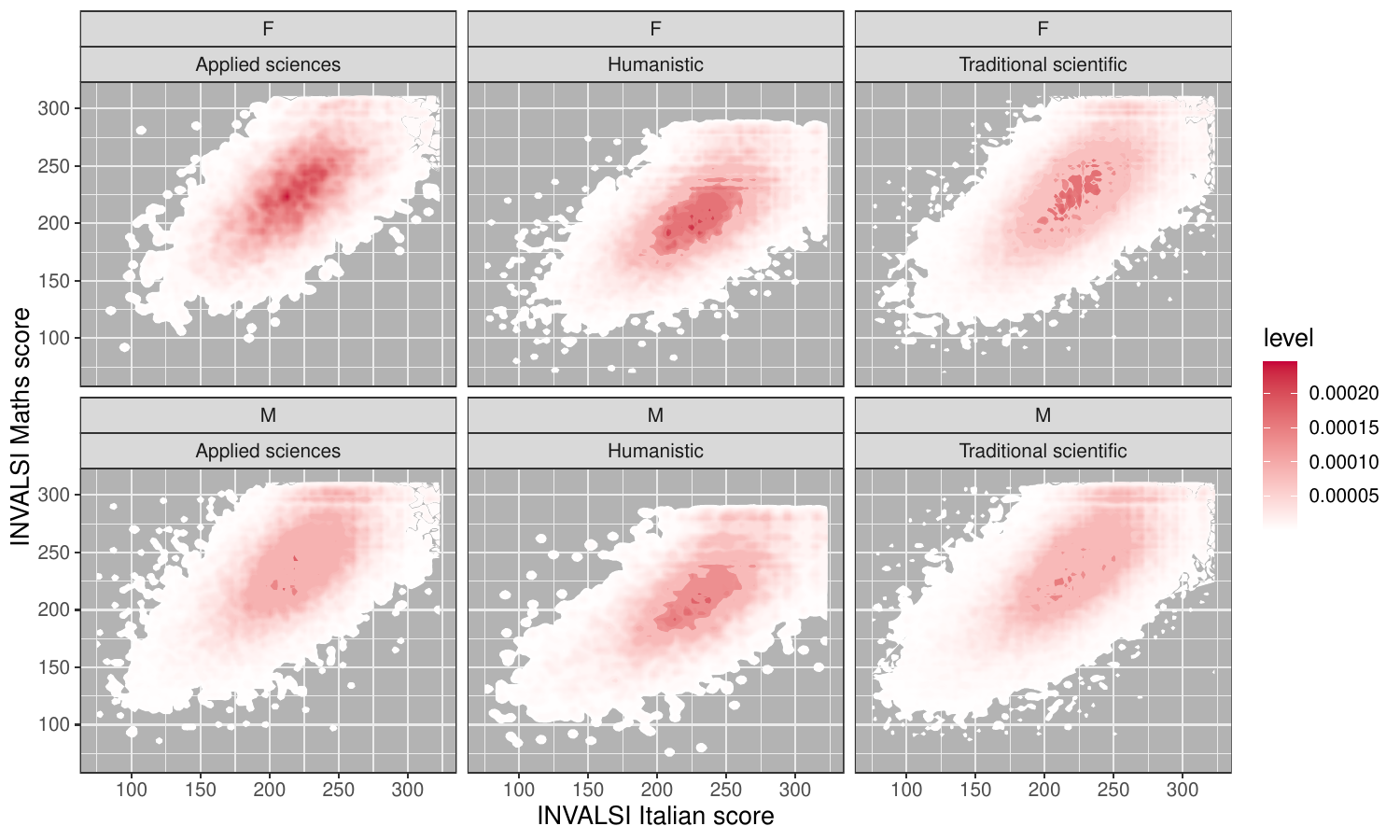}
    \caption{Bivariate distribution of maths and Italian test scores according to gender and high school curriculum.}
        \label{fig:fig3}
\end{figure}

Finally, we show in Figure \ref{fig:fig3} the bivariate joint distributions of maths and Italian test scores, categorized by gender and high school type. Each panel shows a contour plot, where the shapes resemble ellipses following the main bisector. This suggests a positive correlation between math and Italian test scores for both female and male students in each curriculum. In other words, students who perform well in one subject tend to perform well in the other.
Additionally, the plot reveals differences in the densities among the students' profiles. In particular, students from the humanistic curriculum exhibit higher densities, especially females. This implies that the majority of students with this background tend to obtain average scores in both math and Italian tests.

\section{Results}\label{sec:results}
\begin{figure}[h!]
\centering
     \subfloat[][Model 1: University enrolment]{\includegraphics[scale=0.5]{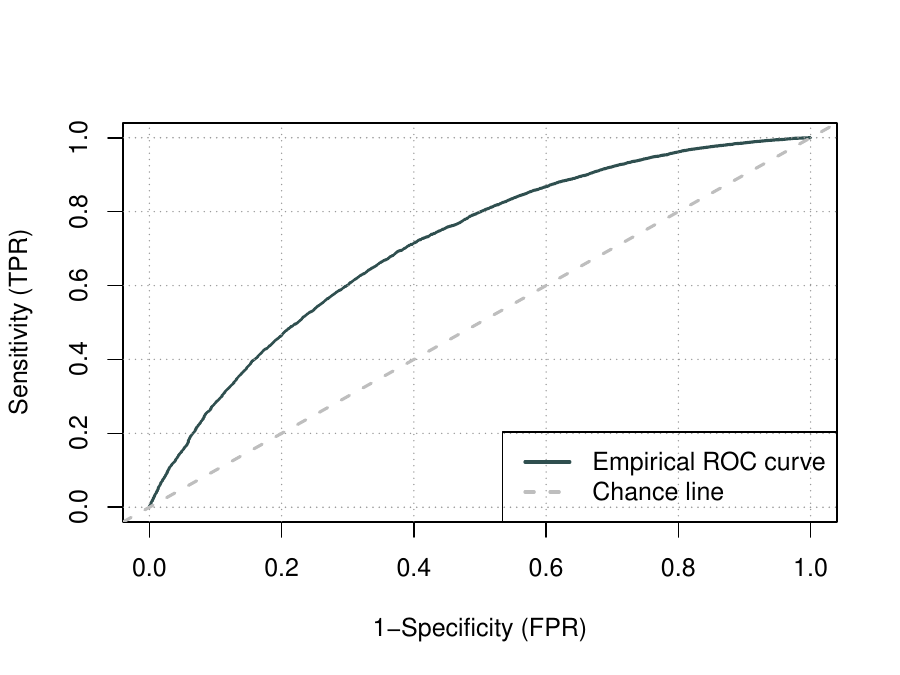}\label{<figure1>}}
     \subfloat[][Model 2: STEM enrolment]{\includegraphics[scale=0.5]{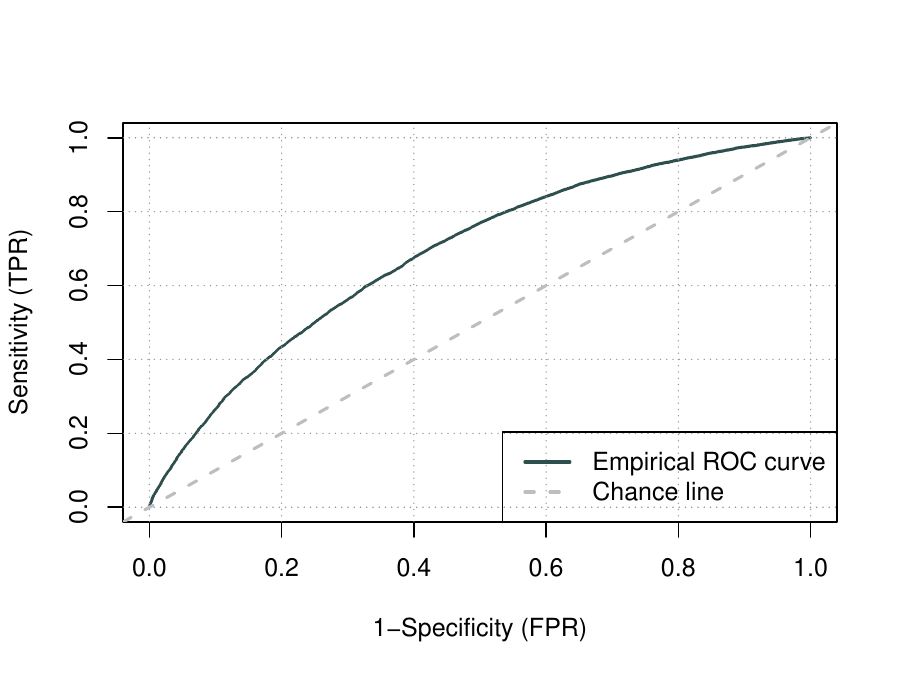}\label{<figure2>}}
   \caption{ROC curves.}
     \label{fig:roc}
\end{figure}

\begin{figure}[h!]
     \centering
     \subfloat[][Model 1: University enrolment]{\includegraphics[scale=0.5]{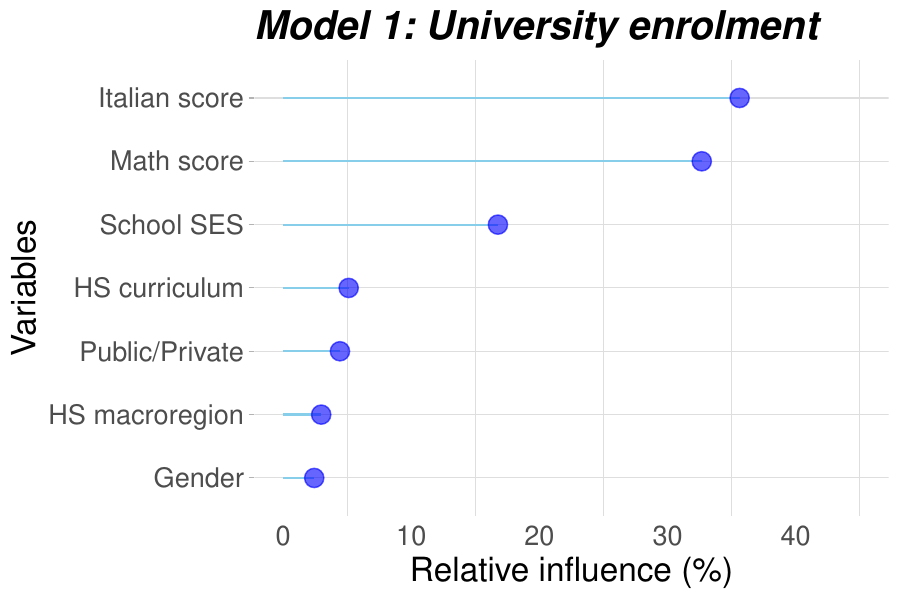}\label{fig:rel1}}
     \subfloat[][Model 2: STEM enrolment]{\includegraphics[scale=0.5]{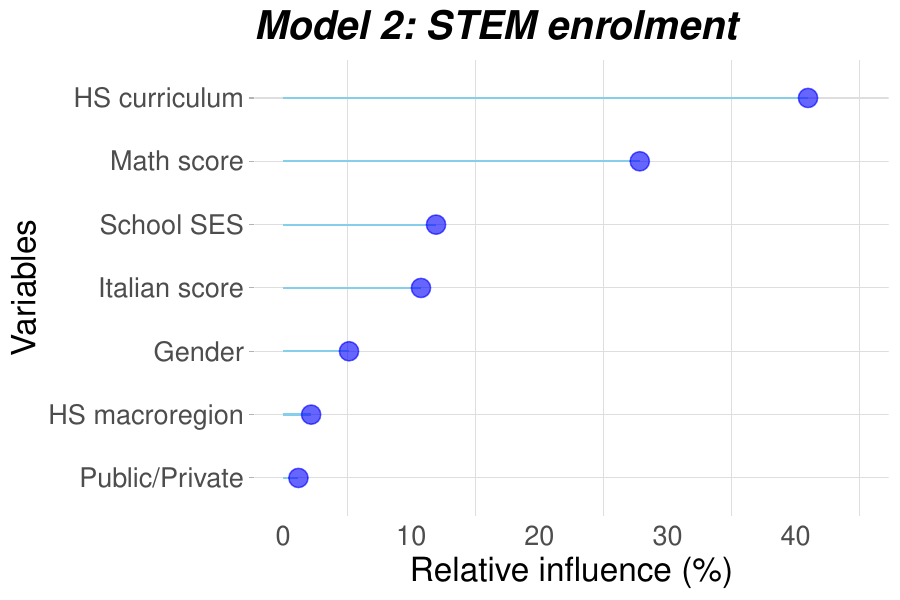}\label{fig:rel2}}
   \caption{Relative influence of predictors.}
     \label{fig:relinf}
\end{figure}

This section presents the key findings derived from our analysis based on GBM. We build two GBM models utilizing the same set of predictors listed in Section \ref{sec:data}. In the first model (named Model 1), we predict the probability of enrollment at an Italian university. Therefore, the reference population for this model comprises all fifth-year high school students in 2018/19. Subsequently, in the second model (Model 2), we shift our focus to predicting enrollment in a STEM program. This model is built on the subset of students who enrolled at an Italian university in 2019/20 as the reference population.
It is important to note that students with an INVALSI math or Italian score of 0 were excluded from the analysis (representing the $0.26\%$ of the total) henceforth. This exclusion is twofold: firstly, based on the assumption that such scores may represent transcription errors; secondly, through sensitivity analysis during the training phase, we observed that including these scores degrades the model's performance.  \\
In both model constructions, the dataset is partitioned into training and test sets, with 75\% of observations allocated to training. The model undergoes hyperparameter tuning on the training set through a 10-fold cross-validation procedure. 
After the final model selection, a thorough evaluation is conducted on the test set, computing the ROC curve and corresponding AUC to assess the model performance comprehensively. 
Figure \ref{fig:roc} provides a visual representation of the ROC curves for the two distinct models on the test sets. 
Both models present well-balanced ROC curves, demonstrating an effective trade-off between sensitivity and specificity. They maintain high sensitivity with a relatively low false-positive rate, underscoring a balanced classification performance.
Model 1 demonstrates robust performance with an AUC of 0.71, while Model 2 exhibits only a slightly less discriminatory capacity, with an AUC of 0.69. \\
We now turn to assessing the influence of predictors in the two models. Firstly, we examine the relative importance of predictors in Figure \ref{fig:relinf}.
In the first model, the most influential variable for predicting university enrolment is the Italian score, accounting for 35.6\% of the relative importance (Figure \ref{fig:rel1}). The math score follows closely at 32.7\%, indicating the significance of high school performance on the university enrolment choice. School SES contributes substantially with 16.8\%, while other factors such as HS curriculum, public/private high school attendance, macroregion, and sex play less significant roles.
It is important to underline that the low relative importance accounted for by the HS curriculum in Model 1 was expected since we considered only those curricula that are more academic-oriented.

In contrast, Figure \ref{fig:rel2} reveals a different pattern of the variables' importance on STEM enrolment. HS curriculum accounts for 41.0\% of the relative importance, highlighting its critical role in predicting STEM enrolment. Math score remains important but takes a secondary position with 27.8\%. School SES, Italian score, and gender follow, contributing to a lesser extent. HS macroregion and Public/Private school attendance exhibit minimal impact in this context, emphasizing the specialized importance of the HS curriculum for STEM outcomes.

\begin{figure}[hbtp!]
     \centering
     \subfloat[]{\includegraphics[scale=0.35]{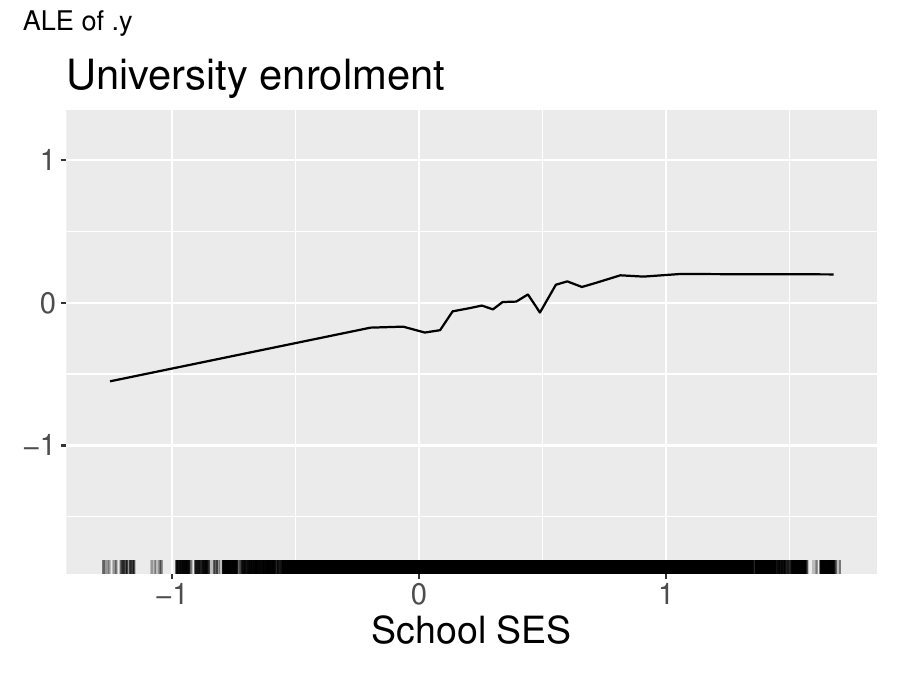}\label{<ale1>}}
     \subfloat[]{\includegraphics[scale=0.35]{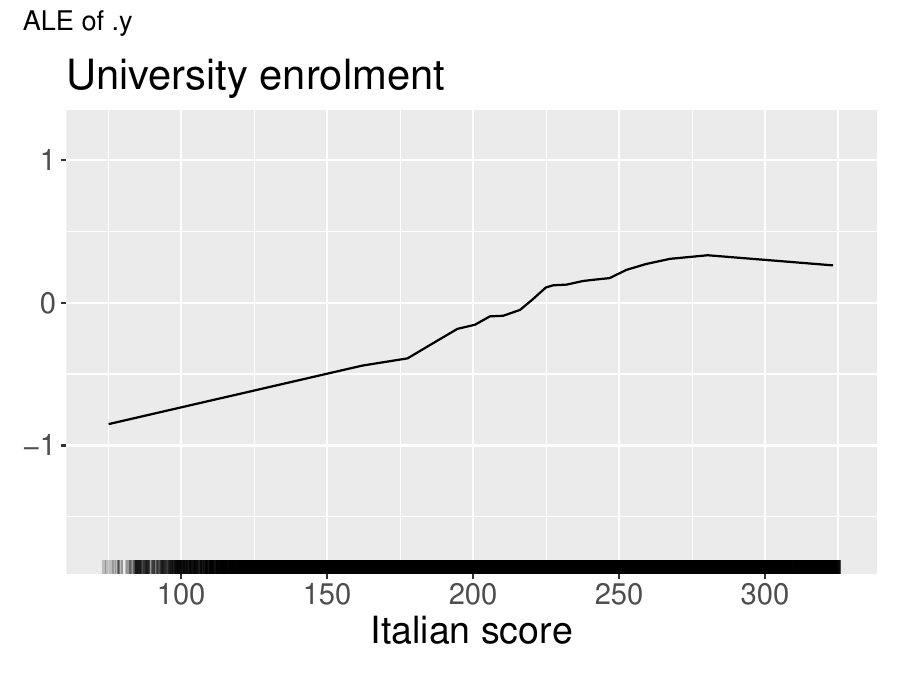}\label{<ale2>}}
     \subfloat[]{\includegraphics[scale=0.35]{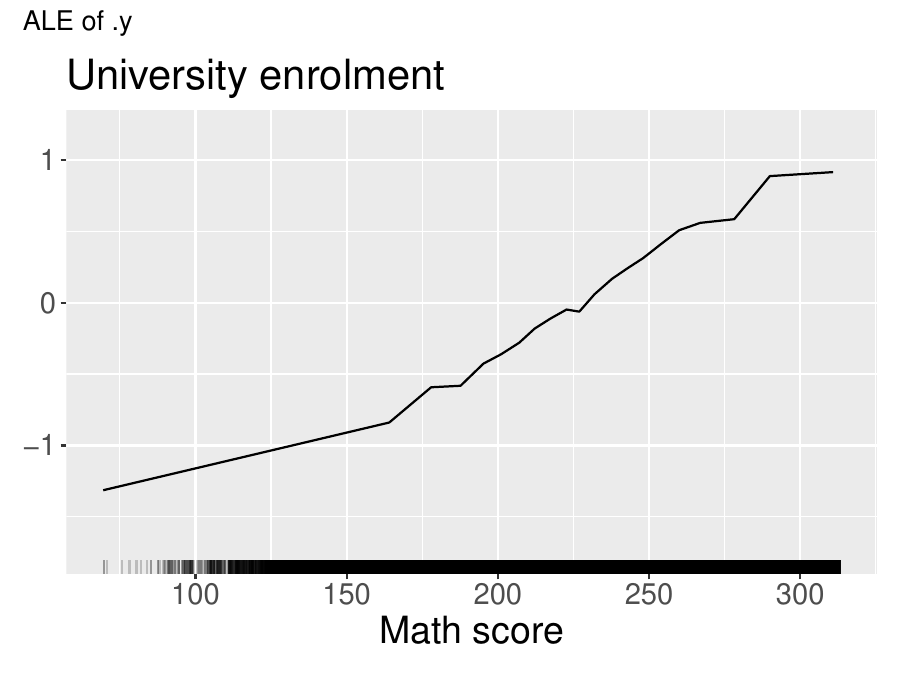}\label{<ale3>}}\\
      \subfloat[]{\includegraphics[scale=0.35]{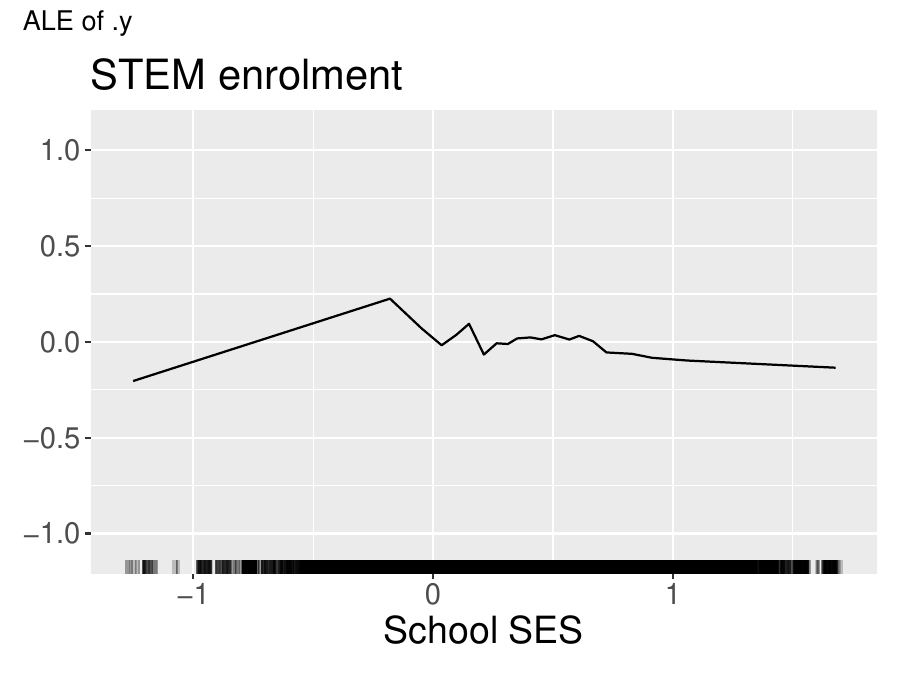}\label{<ale4>}}
     \subfloat[]{\includegraphics[scale=0.35]{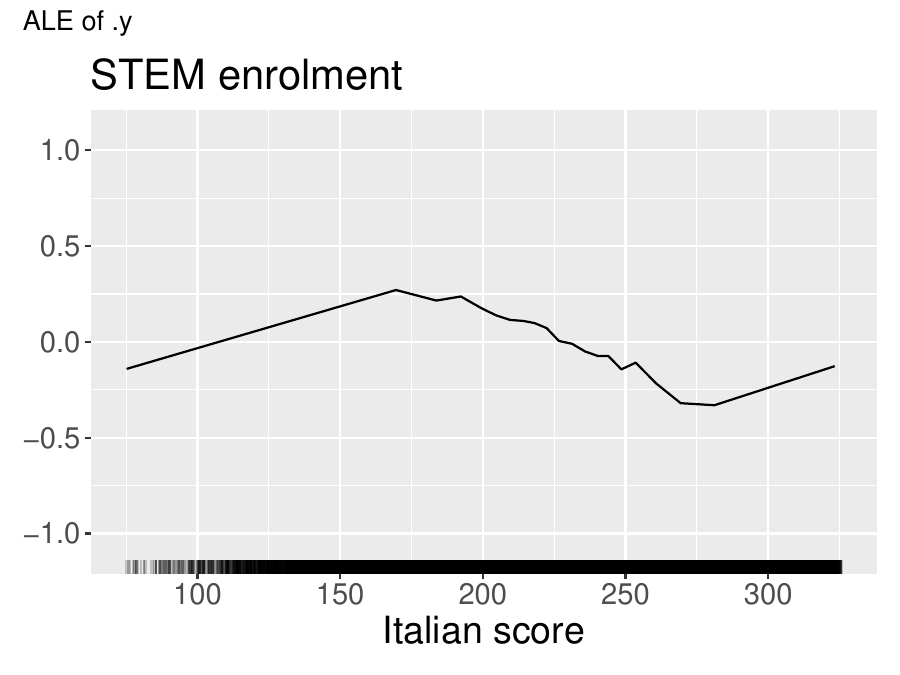}\label{<ale5>}}
     \subfloat[]{\includegraphics[scale=0.35]{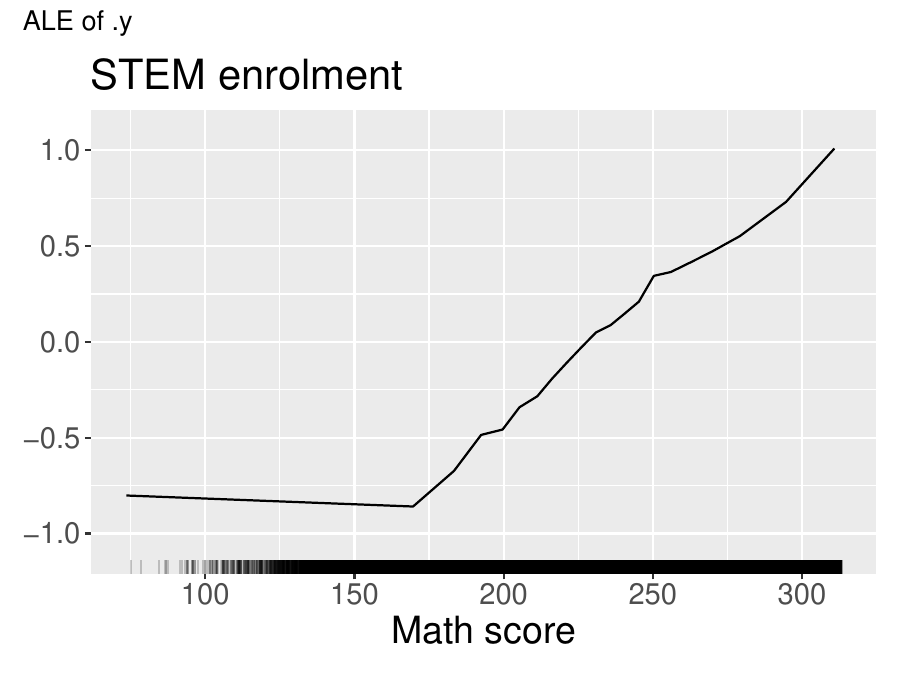}\label{<ale6>}}
     \caption{One-dimensional ALE plots of continuous predictors.}
     \label{fig:Aleplot}
\end{figure}

In Figure \ref{fig:Aleplot}, the ALE (Accumulated Local Effects) of the three continuous explanatory variables in the two models are shown, offering some insights into the nonlinear relationships between probabilities associated with each enrolment choice and the predictors.

In line with the observations from Figure \ref{fig:relinf}, emphasizing the relative importance of variables, it's worth noting that variables with higher importance tend to display more varied patterns in their ALE plots. This underscores the idea that variables of greater influence in the model's predictions often manifest more diverse patterns.

Beginning with the analysis of the socio-economic status of the school attended by the students (Figure \ref{<ale1>}), the graph illustrates a general upward trend, indicating a moderate positive influence on university enrolment in Model 1. This upward trajectory stabilizes after reaching a value of 1. In Model 2, Figure \ref{<ale4>}, the influence of the overall socio-economic status of the school initially rises, dipping slightly below 0, followed by a decreasing trend, ultimately stabilizing for values exceeding 1. 
While in both models, the observed variations are not drastic, they still serve as indicative cues, suggesting that the influence of a very high socio-economic status on university enrolment becomes more nuanced beyond a certain point. Focusing on STEM enrolment, the result indicates that an overall higher school SES negatively influences students' decisions to pursue a STEM career at university. This could be related to the hierarchical structure of the Italian high school system, where students from upper classes are more inclined to attend the humanistic curriculum \citep{panichella2014social}.

Shifting the focus to the Italian score, the analysis reveals a highly nonlinear effect on the responses. In Model 1, Figure \ref{<ale2>}, the influence of the Italian score increases up to a score equal to 250, after which it stabilizes. 
As regards STEM enrolment (Figure \ref{<ale5>}), there is an initial ascending trend up to a score of 175, succeeded by a descending phase up to 275, and then a subsequent ascending trend.  Beyond a critical threshold (175), proficiency in Italian appears to lead to a reduction in STEM enrolments. However, for top performers in Italian, there is once again an ascending trend, indicating that an increase in the score results in more STEM enrolments.

Finally, examining the math score, the ALE analysis unveils an overall positive effect on both the probability of enrolling in any course (Figure \ref{<ale3>}) and the probability of enrolling in a STEM course (Figure \ref{<ale6>}). Particularly notable is the non-linear nature of these increments, especially in Model 2. The ALE exhibits a slightly negative trend between 70 to 170 before undergoing a noticeable sharp increase. It is essential to note that while there is a negative trend during this range, its impact is somewhat negligible, given its almost flat nature. Moreover, particular caution is warranted in the range of 70-100, as the data indicates a scarcity of students in this segment. Consequently, ALE values calculated in this region should be interpreted with due consideration, recognizing the limited sample size and the potential for increased variability in the estimates.

In the following, we shift from ALE to estimated probabilities. This shift aligns with the study's objectives, wherein our primary aim is to investigate the variations in probabilities concerning the two specified academic outcomes. 
To achieve this, we compute multi-dimensional Partial Dependence Plots (PDP), focusing on the interplay of high school curriculum, gender, and performance in the INVALSI Italian and math tests. This method provides a comprehensive understanding of how changes in these variables collectively influence the estimated probabilities of our target outcomes.

\begin{figure}[h!]
    \centering
    \includegraphics[width=\textwidth]{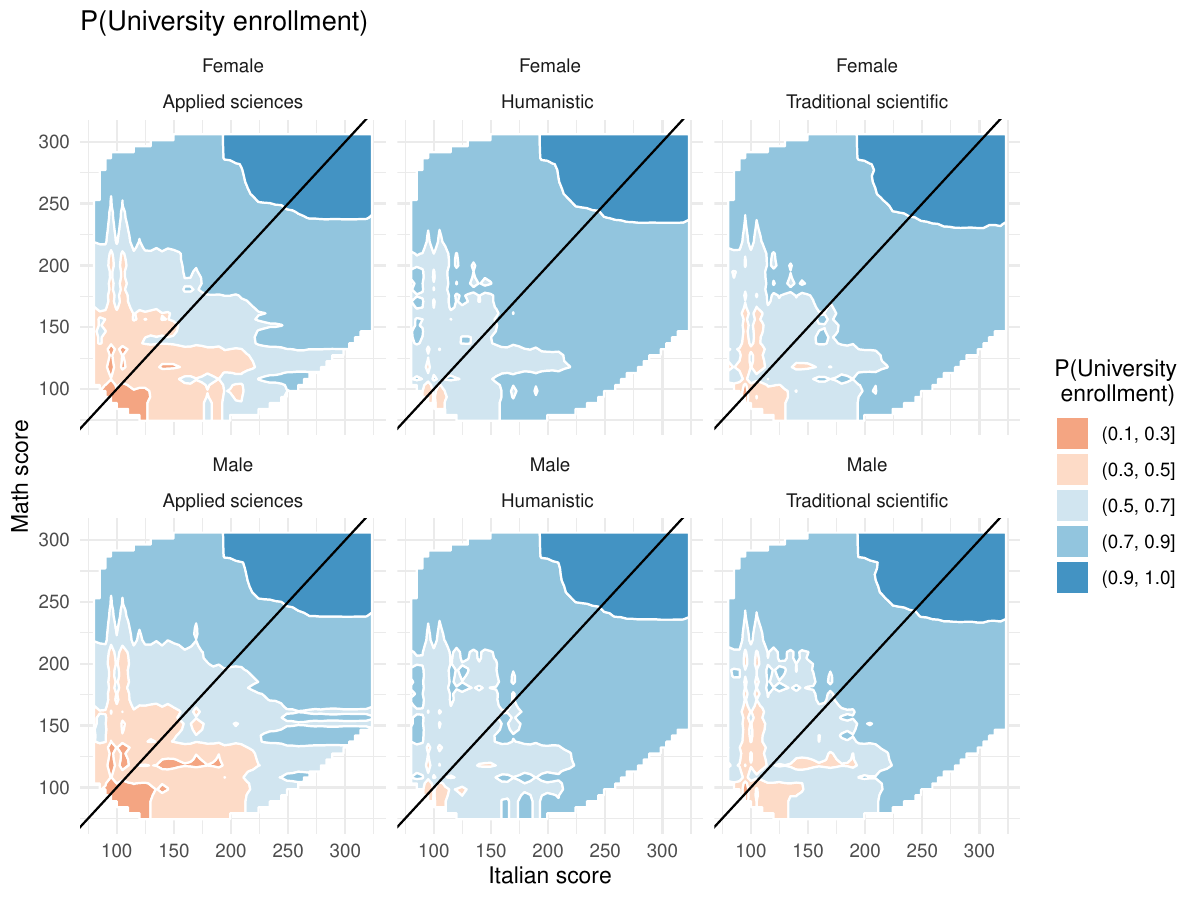}
    \caption{Multidimensional PDP Plot: Probability of enrolling at university based on Italian and math scores, gender and HS curriculum.}
    \label{fig:pred1}
\end{figure}

\begin{figure}[h!]
    \centering
    \includegraphics[width=\textwidth]{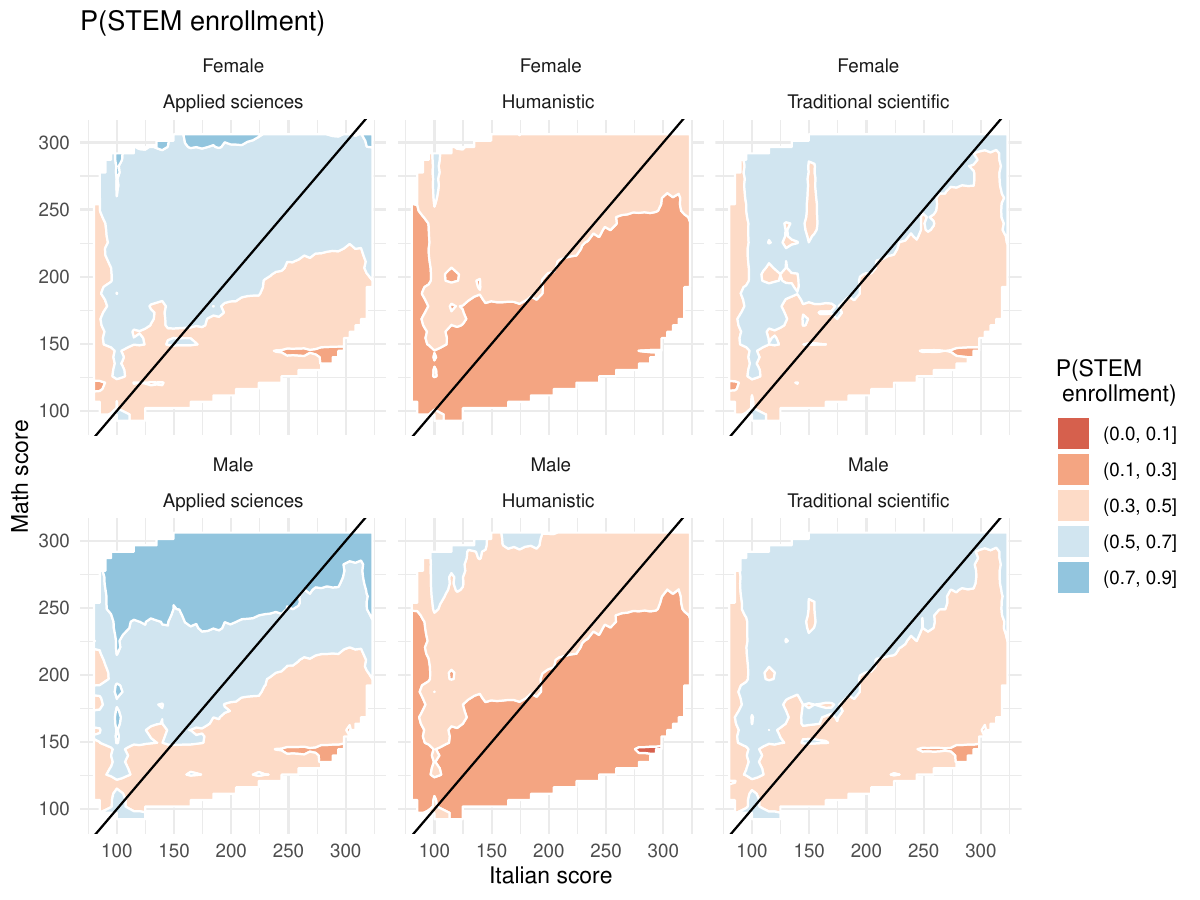}
    \caption{Multidimensional PDP Plot: Probability of enrolling in STEM programs based on Italian and math scores, gender and HS curriculum.}
    \label{fig:pred2}
\end{figure}

Figures \ref{fig:pred1} and \ref{fig:pred2} illustrate the estimated probabilities of university and STEM enrolment, respectively. On the x-axis, we report the Italian score, and on the y-axis, the math scores are shown. 
The three top panels show the estimated probabilities for female students, while the bottom ones show those for males, both conditional on HS curricula. In each panel, the bisector indicates students who achieve the same score on both tests. We will refer to ``top-achieving'' students as those achieving high scores in both tests, namely those placed on the top-right corner of the panels. Conversely, the low-achieving students are found in the bottom-left region of the plots.
As previously said, our analysis primarily focuses on those curricula that train students for a university pathway, more than other HS curricula in Italy. Consequently, the overall probability of university enrolment is notably high (Figure \ref{fig:pred1}). Furthermore, the probability of enrolment increases as scores in both math and Italian tests increase.
Although the estimated probabilities are similar for high-achieving students, some distinctions emerge across the different curricula. Notably, low-achieving students attending the applied sciences curriculum exhibit a lower probability of university enrolment (between 0.1 and 0.3) compared to their low-achieving peers from the other curricula. Specifically, the estimated probabilities for lower-achieving students coming from traditional scientific curriculum range between 0.3 and 0.5, while those with a humanistic background exhibit higher probabilities, ranging between 0.3 and 0.7.\\
Regarding gender differences, male students exhibit a slightly lower propensity to enrol in university compared to their female counterparts. Specifically, poor performance in Italian tests appears to be more effective on the enrollment decisions of male students than on those of female students.
The observed scenario radically changes when it comes to the choice to enrol in a STEM program (Figure \ref{fig:pred2}), where gender and HS curricula exert more influence.
Notably, applied sciences students display a higher inclination to enrol in STEM programs. 
It is also noticeable that the difference between males and females in applied sciences increases as math scores increase. Specifically, male students with a robust math proficiency, scoring above 230, exhibit the highest probability of enrolling in STEM, ranging between 0.7 and 0.9, surpassing their counterparts in other curricula. While female students in applied sciences show a lower probability of enrolling in STEM compared to their male peers, their probability remains notably higher than peers from the humanities and similar to traditional scientific curricula. 
On the other hand, students with a humanities background consistently show lower probabilities of STEM enrollment. Across both panels, the estimated STEM probabilities range between 0.1 and 0.7, with individuals with low math scores falling within the 0.1 to 0.3 probability range. Again, males are more likely to enroll in STEM than females, especially when they exhibit high math scores alongside low Italian scores, reaching a probability range of 0.5 to 0.7. This suggests that, even within the humanities curriculum, a subset of male students with strong quantitative skills demonstrates a noteworthy interest in STEM fields when they perform well in mathematics tests and badly in Italian ones.
Finally, estimated STEM probabilities for scientific curriculum students fall in between those estimated for the other two groups of students, with males exhibiting a slightly higher propensity to enrol in STEM. 


\section{Conclusions}\label{sec:discussion}
This paper has undertaken a comprehensive exploration of the complex relationships between Italian high school students' proficiency in mathematics and in the Italian language, their gender, and their subsequent enrolment choices. The findings reveal compelling insights into the dynamics that shape students' educational choices and contribute to the broader discourse on gender disparities in higher education.

The statistical analysis, conducted using interpretable machine learning methods, specifically gradient boosting, underscores the significant influence of proficiency in both Italian and mathematics on university enrolment decisions. The differentiated impact of these factors on enrolment choices highlights the nuanced nature of students' decision-making processes. Importantly, the study reveals gender differences in enrollment patterns, with male students excelling in both subjects, especially in mathematics, showing a greater inclination toward STEM programs compared to their female counterparts. 
Additionally, the examination of high school backgrounds provides further valuable context, emphasizing the distinctive preferences of male and female students from applied scientific, traditional scientific, and humanistic backgrounds. 

This intersectional approach contributes to a more comprehensive understanding of the interplay between academic performance, gender, and high school background in shaping students' choices. In summary, it becomes evident that students with high math scores are the ones predominantly associated with higher probabilities of enrolling in STEM programs. This trend is particularly pronounced when coupled with low or moderate Italian language scores, and it is more prominent among male students with a scientific background.

Nevertheless, in the intricate landscape of educational choices, these identified variables yet represent only a fraction of the myriad factors that contribute to students' decisions regarding university enrollment and choice of academic programs. The interaction between these variables may be nuanced and subject to contextual variations not captured in this analysis. Factors like individual aspirations, career guidance, and the socio-cultural landscape can also play pivotal roles, contributing to the intricate mosaic of university enrollment dynamics. 
In this framework, this paper also underlines the need for statistical methods that account for the intersectionality of educational data. Intersectionality should be a necessary aspect for scholars to include the diverse interconnections of various dimensions of factors such as socio-economic status, gender, and academic ability, among others.

The paper contributes not only to the academic understanding of educational choices but also holds implications for educational policy. The identified gender disparities and the role of academic proficiency emphasize the need for targeted interventions at the school level. Moreover, the insights into high school backgrounds shed light on the importance of tailoring educational pathways to individual strengths and interests.

Finally, the study has some limitations. In detail, the non-enrolment probability could be slightly overestimated since available data do not provide information about students enrolling at a university abroad. 

\section*{Fundings}
This work has been supported by the Targeted Research Funds 2024 (FFR 2024) of the University of Palermo (Italy).

The research work of Alessandro Albano has been partially supported by the European Union - NextGenerationEU - National Sustainable Mobility Center CN00000023, Italian Ministry of University and Research Decree n. 1033— 17/06/2022, Spoke 2, CUP B73C2200076000.

The research work of Nicoletta D'Angelo has been supported by the European Union -  NextGenerationEU, in the framework of the GRINS -Growing Resilient, INclusive and Sustainable project (GRINS PE00000018 – CUP  C93C22005270001). The views and opinions expressed are solely those of the authors and do not necessarily reflect those of the European Union, nor can the European Union be held responsible for them.

\bibliography{biblio}

\begin{thebibliography}{}

\bibitem[Apley and Zhu, 2020]{apley2020visualizing}
Apley, D.~W. and Zhu, J. (2020).
\newblock Visualizing the effects of predictor variables in black box
  supervised learning models.
\newblock {\em Journal of the Royal Statistical Society Series B: Statistical
  Methodology}, 82(4):1059--1086.

\bibitem[Archer et~al., 2012]{archer2012science}
Archer, L., DeWitt, J., Osborne, J., Dillon, J., Willis, B., and Wong, B.
  (2012).
\newblock Science aspirations, capital, and family habitus: How families shape
  children’s engagement and identification with science.
\newblock {\em American educational research journal}, 49(5):881--908.

\bibitem[Barone and Assirelli, 2020]{barone2020gender}
Barone, C. and Assirelli, G. (2020).
\newblock Gender segregation in higher education: an empirical test of seven
  explanations.
\newblock {\em Higher Education}, 79(1):55--78.

\bibitem[Barone et~al., 2019]{barone2019nudging}
Barone, C., Schizzerotto, A., Assirelli, G., and Abbiati, G. (2019).
\newblock Nudging gender desegregation: A field experiment on the causal effect
  of information barriers on gender inequalities in higher education.
\newblock {\em European Societies}, 21(3):356--377.

\bibitem[Bian et~al., 2017]{bian2017gender}
Bian, L., Leslie, S.-J., and Cimpian, A. (2017).
\newblock Gender stereotypes about intellectual ability emerge early and
  influence children’s interests.
\newblock {\em Science}, 355(6323):389--391.

\bibitem[Borman and Dowling, 2010]{borman2010schools}
Borman, G.~D. and Dowling, M. (2010).
\newblock Schools and inequality: A multilevel analysis of coleman's equality
  of educational opportunity data.
\newblock {\em Teachers College Record}, 112(5):1201--1246.

\bibitem[Boudon, 1974]{boudon1974education}
Boudon, R. (1974).
\newblock Education, opportunity, and social inequality: Changing prospects in
  western society.

\bibitem[Cheryan, 2012]{cheryan2012understanding}
Cheryan, S. (2012).
\newblock Understanding the paradox in math-related fields: Why do some gender
  gaps remain while others do not?
\newblock {\em Sex roles}, 66(3):184--190.

\bibitem[Cheryan et~al., 2017]{cheryan2017some}
Cheryan, S., Ziegler, S.~A., Montoya, A.~K., and Jiang, L. (2017).
\newblock Why are some stem fields more gender-balanced than others?
\newblock {\em Psychological bulletin}, 143(1):1.

\bibitem[Contini et~al., 2023]{contini2023gender}
Contini, D., Di~Tommaso, M.~L., Maccagnan, A., and Mendolia, S. (2023).
\newblock Gender differences in high school choices: Do math and language
  skills play a role?

\bibitem[Contini et~al., 2016]{contini2016between}
Contini, D., Triventi, M., et~al. (2016).
\newblock Between formal openness and stratification in secondary education:
  Implications for social inequalities in italy.
\newblock {\em Models of secondary education and social inequality: An
  international comparison}, pages 305--322.

\bibitem[Correll, 2001]{correll2001gender}
Correll, S.~J. (2001).
\newblock Gender and the career choice process: The role of biased
  self-assessments.
\newblock {\em American journal of Sociology}, 106(6):1691--1730.

\bibitem[Cvencek et~al., 2011]{cvencek2011math}
Cvencek, D., Meltzoff, A.~N., and Greenwald, A.~G. (2011).
\newblock Math--gender stereotypes in elementary school children.
\newblock {\em Child development}, 82(3):766--779.

\bibitem[Delaney and Devereux, 2020]{delaney2020effect}
Delaney, J. and Devereux, P.~J. (2020).
\newblock The effect of high school rank in english and math on college major
  choice.

\bibitem[Fernandes et~al., 2019]{fernandes2019educational}
Fernandes, E., Holanda, M., Victorino, M., Borges, V., Carvalho, R., and
  Van~Erven, G. (2019).
\newblock Educational data mining: Predictive analysis of academic performance
  of public school students in the capital of brazil.
\newblock {\em Journal of business research}, 94:335--343.

\bibitem[Friedman et~al., 2000]{friedman2000additive}
Friedman, J., Hastie, T., and Tibshirani, R. (2000).
\newblock Additive logistic regression: a statistical view of boosting (with
  discussion and a rejoinder by the authors).
\newblock {\em The annals of statistics}, 28(2):337--407.

\bibitem[Friedman, 2001]{friedman2001greedy}
Friedman, J.~H. (2001).
\newblock Greedy function approximation: a gradient boosting machine.
\newblock {\em Annals of statistics}, pages 1189--1232.

\bibitem[Friedman, 2002]{friedman2002stochastic}
Friedman, J.~H. (2002).
\newblock Stochastic gradient boosting.
\newblock {\em Computational statistics \& data analysis}, 38(4):367--378.

\bibitem[Friedman-Sokuler and Justman, 2016]{friedman2016gender}
Friedman-Sokuler, N. and Justman, M. (2016).
\newblock Gender streaming and prior achievement in high school science and
  mathematics.
\newblock {\em Economics of Education Review}, 53:230--253.

\bibitem[Gabay-Egozi et~al., 2015]{gabay2015gender}
Gabay-Egozi, L., Shavit, Y., and Yaish, M. (2015).
\newblock Gender differences in fields of study: the role of significant others
  and rational choice motivations.
\newblock {\em European Sociological Review}, 31(3):284--297.

\bibitem[Giambona and Porcu, 2018]{giambona2018school}
Giambona, F. and Porcu, M. (2018).
\newblock School size and students' achievement. empirical evidences from pisa
  survey data.
\newblock {\em Socio-Economic Planning Sciences}, 64:66--77.

\bibitem[Gonz{\'a}lez-P{\'e}rez et~al., 2020]{gonzalez2020girls}
Gonz{\'a}lez-P{\'e}rez, S., Mateos~de Cabo, R., and S{\'a}inz, M. (2020).
\newblock Girls in stem: Is it a female role-model thing?
\newblock {\em Frontiers in psychology}, 11:2204.

\bibitem[Gr{\"o}mping, 2020]{gromping2020model}
Gr{\"o}mping, U. (2020).
\newblock Model-agnostic effects plots for interpreting machine learning
  models.
\newblock {\em Reports in Mathematics, Physics and Chemistry, Department II,
  Beuth University of Applied Sciences Berlin Report}, 1:2020.

\bibitem[Hadjar, 2019]{hadjar2019educational}
Hadjar, A. (2019).
\newblock Educational expansion and inequalities: how did inequalities by
  social origin and gender decrease in modern industrial societies.
\newblock {\em Research handbook on the sociology of education. Cheltenham, UK:
  Edward Elgar Publishing}, pages 173--192.

\bibitem[Hadjar and Becker, 2009]{hadjar2009}
Hadjar, A. and Becker, R. (2009).
\newblock Educational expansion: Expected and unexpected consequences.
\newblock {\em Expected and Unexpected Consequences of the Educational
  Expansion in Europe and the US. Bern: Haupt}, pages 9--23.

\bibitem[Hadjar and Buchmann, 2016]{hadjar2016education}
Hadjar, A. and Buchmann, C. (2016).
\newblock Education systems and gender inequalities in educational attainment.
\newblock In {\em Education systems and inequalities}, pages 159--184. Policy
  Press.

\bibitem[Hilbert et~al., 2021]{hilbert2021machine}
Hilbert, S., Coors, S., Kraus, E., Bischl, B., Lindl, A., Frei, M., Wild, J.,
  Krauss, S., Goretzko, D., and Stachl, C. (2021).
\newblock Machine learning for the educational sciences.
\newblock {\em Review of Education}, 9(3):e3310.

\bibitem[{ISTAT}, 2023]{ISTAT2023}
{ISTAT} (2023).
\newblock Rapporto {BES} 2022, ``il benessere equo e sostenibile in italia'''.

\bibitem[Kromydas, 2017]{kromydas2017}
Kromydas, T. (2017).
\newblock Rethinking higher education and its relationship with social
  inequalities: past knowledge, present state and future potential.
\newblock {\em Palgrave communications}, 3(1):1--12.

\bibitem[Legewie and DiPrete, 2012]{legewie2012school}
Legewie, J. and DiPrete, T.~A. (2012).
\newblock School context and the gender gap in educational achievement.
\newblock {\em American Sociological Review}, 77(3):463--485.

\bibitem[L{\"o}rz and M{\"u}hleck, 2019]{lorz2019gender}
L{\"o}rz, M. and M{\"u}hleck, K. (2019).
\newblock Gender differences in higher education from a life course
  perspective: transitions and social inequality between enrolment and first
  post-doc position.
\newblock {\em Higher Education}, 77:381--402.

\bibitem[Macarie and Moldovan, 2015]{macarie2015horizontal}
Macarie, F.~C. and Moldovan, O. (2015).
\newblock Horizontal and vertical gender segregation in higher education: Eu 28
  under scrutiny.
\newblock {\em Managerial Challenges of the Contemporary Society. Proceedings},
  8(1):162.

\bibitem[Makarova et~al., 2019]{makarova2019gender}
Makarova, E., Aeschlimann, B., and Herzog, W. (2019).
\newblock The gender gap in stem fields: The impact of the gender stereotype of
  math and science on secondary students' career aspirations.
\newblock In {\em Frontiers in Education}, page~60. Frontiers.

\bibitem[McNally, 2020]{mcnally2020gender}
McNally, S. (2020).
\newblock Gender differences in tertiary education: what explains stem
  participation?
\newblock Technical report, IZA Policy Paper.

\bibitem[MOBYSU.IT, 2017]{mobysu2017}
MOBYSU.IT (2017).
\newblock {\em Database MOBYSU.IT, Mobilit{\`a} degli studi universitari
  italiani, Research Protocol MUR - Universities of Cagliari, Palermo, Siena,
  Torino, Sassari, Firenze, Cattolica and Napoli Federico II, Scientific
  Coordinator Massimo Attanasio (UNIPA), Data Source ANS-MUR/CINECA}.

\bibitem[Molnar, 2020]{molnar2020interpretable}
Molnar, C. (2020).
\newblock {\em Interpretable machine learning}.
\newblock Lulu. com.

\bibitem[OECD, 2022]{oecd2022}
OECD (2022).
\newblock {\em Education at a Glance 2022}.

\bibitem[Panichella and Triventi, 2014]{panichella2014social}
Panichella, N. and Triventi, M. (2014).
\newblock Social inequalities in the choice of secondary school: Long-term
  trends during educational expansion and reforms in italy.
\newblock {\em European Societies}, 16(5):666--693.

\bibitem[Porcu et~al., 2022]{porcu2022estimating}
Porcu, M., Sulis, I., Usala, C., et~al. (2022).
\newblock Estimating the peers effect on students' university choices.
\newblock In {\em Book of short papers. IES 2022 Innovation \& society 5.0:
  statistical and economic methodologies for quality assessment}, pages
  134--139. PKE srl.

\bibitem[Priulla and Attanasio, 2023]{priulla2023unveiling}
Priulla, A. and Attanasio, M. (2023).
\newblock Unveiling gender disparities in university pathways: insights from
  italy’s master’s level.
\newblock {\em European Journal of Higher Education}, pages 1--24.

\bibitem[Priulla et~al., 2023]{priulla2023does}
Priulla, A., Vittorietti, M., and Attanasio, M. (2023).
\newblock Does taking additional maths classes in high school affect academic
  outcomes?
\newblock {\em Socio-Economic Planning Sciences}, page 101674.

\bibitem[Regan and DeWitt, 2014]{regan2014attitudes}
Regan, E. and DeWitt, J. (2014).
\newblock Attitudes, interest and factors influencing stem enrolment behaviour:
  An overview of relevant literature.
\newblock {\em Understanding student participation and choice in science and
  technology education}, pages 63--88.

\bibitem[Ridgeway, 2007]{ridgeway2007generalized}
Ridgeway, G. (2007).
\newblock Generalized boosted models: A guide to the gbm package.
\newblock {\em Update}, 1(1):2007.

\bibitem[Romito et~al., 2020]{romitogender}
Romito, M., Gerosa, T., Visentin, M., and Maria, G. (2020).
\newblock Gender differences in higher education choices. italian girls in the
  corner?
\newblock {\em The Education of Gender The Gender of Education}, page~61.

\bibitem[Salmieri, 2022]{salmieri2022students}
Salmieri, L. (2022).
\newblock Students, parents and school-choices. gendered trajectories in the
  italian education system.
\newblock {\em Italian Journal of Sociology of Education}, 14(2):99--119.

\bibitem[Salmieri and Giancola, 2020]{salmieri2020gender}
Salmieri, L. and Giancola, O. (2020).
\newblock Gender differences in tertiary educational attainment and the
  intergenerational transmission of cultural capital in italy.
\newblock {\em The Education of Gender The Gender of Education}, page~77.

\bibitem[Sherman, 1980]{sherman1980mathematics}
Sherman, J. (1980).
\newblock Mathematics, spatial visualization, and related factors: Changes in
  girls and boys, grades 8--11.
\newblock {\em Journal of Educational psychology}, 72(4):476.

\bibitem[Smith and Gorard, 2011]{smith2011there}
Smith, E. and Gorard, S. (2011).
\newblock Is there a shortage of scientists? a re-analysis of supply for the
  uk.
\newblock {\em British Journal of Educational Studies}, 59(2):159--177.

\bibitem[Stoet and Geary, 2018]{stoet2018gender}
Stoet, G. and Geary, D.~C. (2018).
\newblock The gender-equality paradox in science, technology, engineering, and
  mathematics education.
\newblock {\em Psychological science}, 29(4):581--593.

\bibitem[Suzuki et~al., 2022]{suzuki2022prediction}
Suzuki, H., Hong, M., Ober, T., and Cheng, Y. (2022).
\newblock Prediction of differential performance between advanced placement
  exam scores and class grades using machine learning.
\newblock In {\em Frontiers in Education}, volume~7, page 1007779. Frontiers.

\bibitem[Tandrayen-Ragoobur and Gokulsing, 2021]{tandrayen2021gender}
Tandrayen-Ragoobur, V. and Gokulsing, D. (2021).
\newblock Gender gap in stem education and career choices: what matters?
\newblock {\em Journal of Applied Research in Higher Education}.

\bibitem[Tey et~al., 2020]{tey2020teacher}
Tey, T. C.~Y., Moses, P., and Cheah, P.~K. (2020).
\newblock Teacher, parental and friend influences on stem interest and career
  choice intention.
\newblock {\em Issues in Educational Research}, 30(4):1558--1575.

\bibitem[Ya{\u{g}}c{\i}, 2022]{yaugci2022educational}
Ya{\u{g}}c{\i}, M. (2022).
\newblock Educational data mining: prediction of students' academic performance
  using machine learning algorithms.
\newblock {\em Smart Learning Environments}, 9(1):11.

\end{thebibliography}

\end{document}